\documentclass[letterpaper,twocolumn,fleqn]{article} 

\usepackage{ist}
\usepackage{comment}
\usepackage{bm}

\title{A Simple and efficient deep Scanpath Prediction}

\author{ Mohamed Amine KERKOURI,  Aladine CHETOUANI\\
Laboratoire PRISME, Université d'Orléans, Orléans, FRANCE }

\date{} % date has an empty field.

\begin{document} 

\maketitle 

\thispagestyle{empty} % prevents the first page to be numbered

%%%%%%%%%%%%%%%%%%%%%%%%%%%%%%%%%%
% Abstract
%%%%%%%%%%%%%%%%%%%%%%%%%%%%%%%%%%

\begin{abstract}
Visual scanpath is the sequence of fixation points that the human gaze travels while observing an image, and its prediction helps in modeling the visual attention of an image. To this end several models were proposed in the literature using complex deep learning architectures and frameworks. Here, we explore the efficiency of using common deep learning architectures, in a simple fully convolutional regressive manner. We experiment how well these models can predict the scanpaths on 2 datasets. We compare with other models using different metrics and show competitive results that sometimes surpass previous complex architectures. We also compare the different leveraged backbone architectures based on their performances on the experiment to deduce which ones are the most suitable for the task.  
\end{abstract}

%%%%%%%%%%%%%%%%%%%%%%%%%%%%%%%%%%%%
% Overall Document Guidelines: Head
%%%%%%%%%%%%%%%%%%%%%%%%%%%%%%%%%%%%
\section{1. Introduction}
%\label{sec:introduction}
%\PARstart

% human visual system complexity and efficiency
% research to memic the HVS ( computer vision)  
% visual attention a very important characteristic 
% What is visual attention
% Use cases of visual attention
% datasets
% Saliency vs scanpath prediction
% importance of scanpath prediction 
%  work of paper ( introduction of net., comparison )
% paper structure 
% 

The Human Visual System (HSV) is one of the most advanced and efficient recognition systems that inspired a lot of researchers for different visual tasks. Visual attention is among the prominent characteristics of this system. It is a mechanism that tends to consistently focus on specific regions of an image, decreasing the  computational load during the visual task. The choice of these regions is influenced by low level characteristics of the image (Color, texture, intensity, ... etc.) and high level features (presence of faces, objects, ... etc.) \cite{feature_integration}, this is called  "Bottom-Up" attention, as it is driven by the low level features of the image. The attention can also be influenced by higher cognitive functions (search for a specific object, ... etc), where prior information is available, and a specific task is guiding the gaze. This is "Top-down" attention. These attractive areas are called salient regions, and are represented using 2D probabilistic heatmaps called saliency maps. The value of each pixel estimates its attractiveness. The saliency maps are constructed by measuring the density of the points of gaze fixation of several observers. For each observer, the sequence of fixation points travelled by his gaze is called a scanpath.

%which are estimated using by the density of fixation points of the 

%They are estimated by the capturing of eye movement during the observation using eye-trackers\cite{Judd}, mouse clicks \cite{Salicon} or webcams \cite{turckergaze}. 
%This is called the overt attention, as it can be observed and measured, another class of visual attention is the covert attention which can not be measured,as it takes place in higher cognitive level in the mind. 

%The attention process is driven by several factors like the features of the images (Color, intensity, orientation, semantics of the image), this called a "Bottom-Up" attention as it's driven by the low level features of the image. The attention can also be influenced by higher cognitive functions like    search for a specific object, where prior information is available, this is "Top-down" attention.      

%The sequence of points estimated create the human visual saccadic scanpath. Points from several observers are aggregated to gray scale heat-map describing the probability distribution of the attractiveness of the regions, these maps are called the saliency map. 

Saliency prediction is used to improve several applications like subjective image quality assessment \cite{QAChetouaniICIP2018}, image and video compression \cite{saliencyComp1}, image captioning and description \cite{saliencyCapt1}, image search and retrieval \cite{retrivalSal}.
% Related works 
% heuristic
Several methods have been proposed, starting with the seminal work of \cite{KochUllman}, later implemented by Itti et al. \cite{Itti} in hierarchical multi-scale model that uses low level features (i.e. color, intensity and orientation) to construct a saliency map.  Harel et al.   \cite{GBVS}, proposed to predict the saliency based on the graph theory where Markov chains is defined over different input maps. While Bruce et al. \cite{infosal} proposed a model based on information maximization using the Shannon self-information measure. Other studies like \cite{spectral1} and \cite{spectral2} explored the saliency prediction using spectral domain of the images. In recent years, the introduction of deep learning models has increased the precision of the results in multiple computer vision areas, and they were used in several works for saliency prediction. One of the first methods has been proposed by Pan et al.\cite{PanDSN} where a deep and a shallow neural network are used for saliency prediction. Kummerer et al. \cite{deepgaze1} introduced DeepGaze I where they trained \cite{deepgaze1ref} network on MIT dataset \cite{Judd} for saliency map prediction. They later introduced another model by using a VGG-19 as baseline, improving his results. Cornia et al. \cite{mlnet} used a Multi-level feature Network. They extracted feature maps from different levels on the network and concatenated them. The resulting feature maps were passed through a convolutional layer to predict saliency maps. Jiang et al. \cite{Salicon} used a multi scale approach, where 2 CNN networks trained on different image scales and merged the results to predict the saliency maps. Pan et al. \cite{salgan} introduced the used of generative modeling and adversarial learning for the proposed SalGan Model, by using a deep convolutional Generative Adversarial Network (GAN). While Cornea et al. \cite{sam} combined convolutional model with recurrent Long Short Term Memory (LSTM) to introduce an attention mechanism. In 2020, Drost et al. \cite{unisal} proposed (Unisal) a unified model for saliency prediction, in both static images and dynamic video.

%While we presented quite a number of works on saliency prediction, it was not a full review of all  the works done on the subject. The available literature on scanpath prediction task is much smaller, because the interest in it has risen only this past few years.   
The research concerning the prediction of scanpaths, got more attention lately, but it started by the use of winner-takes-all approach in \cite{Itti}. Later \cite{boccignone} modeled the human saccadic process as a random walk with a jump of stochastic length and angle. In 2007, Cerf et al. \cite{cerf} discovered that observers, in free viewing task, fixate on a human faces with a probability of over $80\%$ within their first two fixations while publishing the Fixations in Faces (FiFA) database. A stochastic model where visual fixations are inferred from saliency maps and oculomotor biases (i.e. saccade amplitudes and saccade orientations) calculated from several datasets was proposed by Le meur et al. \cite{lemeur}. While in \cite{G-Eymol} the authors compared the gaze to traveling mass point in the image space and the salient regions to gravitational fields affecting speed and trajectory of the mass. Shao et al. \cite{shao} proposed a model that uses High-Level Features from CNN and Memory Bias including short-term and long-term memory for scanpath prediction. Assens et al. \cite{saltinet} proposed Saltinet, a deep learning model that infers the saccades from a static saliency volume generated by an encoder-decoder network. The inference is here done using different sampling strategies. To increase the dependence between scanpath fixations, they then proposed a second model PathGAN \cite{pathgan} that uses LSTM Layers to infer scanpaths from a VGG extracted feature maps , they used a conditional adversarial learning approach in order to increase the accuracy of their predictions. Verma et al. \cite{HMMLSTMConv} used an LSTM combined with convolutional network to predict the scanpaths. They employed an HMM (Hidden Markov Model) to augment the data for the LSTM. While Bao et al. \cite{DCSM} proposed a framework called Deep Convolutional Saccadic Model (DCSM) which predicted the fovial saliency maps and temporal duration while modeling the Inhibition of Return (IoR). Kerkouri et al. \cite{salypath} proposed a model which simultaneously predicts saliency and scanpaths from  the same latent feature.   

In this paper we :  
\begin{itemize}
  \item Use regression with a fully convolutional network to  predict plausible scanpaths for natural images.
  \item Compare the suitability of different well known CNN architectures for the task.   
  \item Achieve results comparable to the state-of-the-art using a simple Mean Squared Error (MSE) loss function on multiple datasets. 
  \item Compare the efficiency of our approach to previous models.      

\end{itemize}

%The rest of the paper will review the previous work for both saliency and scanpath prediction in section 2. Section 3 will introduce the our proposed method, the datasets used and training process.  
%In section 4 we discuss the results obtained, and finalise with a conclusion in  section 5. 

%Jiang et al. \cite{LSPI} used LSPI (least-squares policy iteration) a reinforcement learning technique to simulate the the saccadic movement of the human eye.They suggested that the low-level and high-level features  are similarly important at the beginning of the scanpath, but the importance of high-level features increases during the visual exploration of the image.

%Shao et al. \cite{shao}  proposed a model that uses  High-Level Features from CNN and Memory Bias  including short-term and long-term memory for scanpath prediction.  

%Kummerer et al. \cite{extdeepgaze} extended their previous model for saliency prediction DeepGaze II to the human scanpath prediction task by simply adding the features to encode the previous scanpath.

The rest of this paper presents the used methods, datasets and training process in Section 2. Sections 3 and 4 describe the testing protocol and discuss the obtained results, respectively. Finally, we conclude in Section 5.

\section{2. Proposed method}
\label{method}

The goal of this study is to propose a simple method that efficiently predicts a scanpath of an image, and compare it with more complex methods.  %We first briefly describe the datasets used in our experiments. 
We first present the general architecture of the proposed framework and the details of its design. Then, we describe the training protocol applied for the models of our framework.

\subsection{2.1. Architecture}

\begin{figure}
    \vskip 0.5cm
    \includegraphics[width=\linewidth]{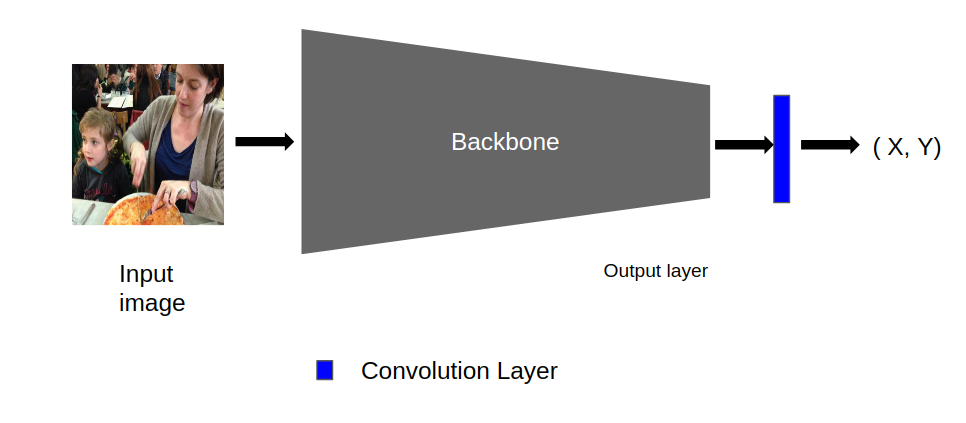}
    \caption{\label{fig:general_arch} General architecture of our model.}
    \vspace{-6mm}
\end{figure}

The general architecture of the proposed scanpath predictor is illustrated in Fig. \ref{fig:general_arch}. For a given image, we first extract feature maps from one of the chosen backbone CNN architecture. These maps are then fed to a further readout convolution layer with an appropriate kernel size in order to predict a scanpath of determined size for the image. 
%Using a fully convolutional network benefits from GPU acceleration and parallelization during training, unlike LSTM based models like PathGan
%It is worth noting that unlike the PathGAN, our model can benefit from the GPU acceleration and parallelization during training, this leads to faster and less memory consuming training. 

\subsubsection{2.2.1 Backbone}

The backbone is used to extract high level features from the images. These features will be used by the readout convolutional layer in order to predict a scanpath. In this study, we used 4 network architectures of well famous  models and compare their efficiency for this purpose. The following is a description of the models for these backbone architectures :  
%Here, 4 pre-trained CNN models have been compared on both datasets. Each model has been widely used computer vision: 
\begin{itemize}
    \item  VGG16\cite{vgg} : It introduced the use of a sequence of blocks. They were composed of a succession of convolutional layers followed by max pooling. The architecture introduced a simple yet deep architecture, %expanding the receptive fields compared to the previous architectures. 
    able to learn higher level features compared to previous models. %It achieved competitive results in ILSVRC 2014 competition \cite{imagenet_org} \cite{imagenet2015}. 
    In our model the output feature maps of the layer usually designated as \textit{block5\_conv3}\footnote{The names used for layer are according models in the Keras library} is selected and passed to the readout convolution layer. 
    \item  ResNet50 \cite{resnet}: It presented the concept of residual blocks and skip connections. For each block, the input feature maps are added to its output. This solution was used to resolve the vanishing gradient problem for deep architectures. In our model, The output feature maps of the layer \textit{conv4\_block23\_out} is selected and passed to the added convolution layer.  % (9x15) 
    \item  InceptionV3 \cite{inception}: It proposed the inception module as a building block that employs parallel convolutional layers with different kernel sizes. The InceptionV2 added the factorization of the kernels into (1xn) and (nx1) kernel sequences, while the InceptionV3 added a factorized 7x7 kernel among other contributions. The output of the layer \textit{activation\_74} in the 9\textsuperscript{th} inception block is selected and passed to the added convolution layer.  % ( 7x 13 ) 
    \item  DenseNet121 \cite{densenet}: It introduced blocks that use high density of long and short skip connections. The output of each layer in the block is propagated to all the following layers in the same block. Unlike ResNet \cite{resnet} which adds the input to the output, DenseNet \cite{densenet} uses concatenation instead which led to a big improvement in results. The output of the layer \textit{pool4\_conv}, which is located after the 24\textsuperscript{th} dense block, is selected and passed to the added convolution layer.\\ 
\end{itemize}
%\useshortskip
%\vspace{-5mm} 
For the rest of this work, we will refer to the models as \{backbone name\}-BB (e.g VGG-BB, ...).   

\subsubsection{2.2.2. Readout Convolution Layer }

In order to determine the optimal length of scanpath and thus the size of the kernel of the readout convolution layer, a statistical analysis was conducted on the lengths of scanpaths on Salicon dataset \cite{Salicon}. Table 1 presents the results obtained. The statistical central tendency indicates that a length of 8 fixation points is appropriate, while the dispersion tendency measures, the standard deviation of 4.45 compared to the range of 35 reinforces this decision. 

\begin{table*}[ht]
\begin{center}
\begin{tabular}{ c  c c c c c c c c}
\hline
\textbf{Measure} &  Min & Max & Mean & Median & std & Mode &   Nbr. scanpaths \\
\hline
 \textbf{Value} & 1 & 35 & 7.86 & 8 & 4.45 & 8 (9.38 \%) & 584927  \\ 
 \hline
 
\end{tabular}
\label{tab:Salicon_analysis}
\caption{Table 1. Statistical analysis on Salicon.}
\end{center}
\vspace{-6mm}
\end{table*}

Each of the considered models takes an \textit{224x224x3} image as input and outputs an 8x2 feature map, representing the spatial coordinates of the 8 predicted fixations of the scanpath. Therefore, a kernel of size $(9x15)$ was chosen for VGG-BB, ResNet-BB and DenseNet-BB for the readout convolution layer, while a size of $(7x13)$ was chosen for Inception-BB. It is important to mention here that the temporal coordinates of the saccades is not considered.  

\subsection{2.3. Training}

All of our models were trained using the MSE Loss: 
\[Loss=\frac{1}{N} \sum_{i=1}^{N} (y_i-\hat{y_i})^2 \]
where $y_i$ represents the predicted scanpath and $\hat{y_i}$ represents the ground truth scanpath. 

\begin{figure}
    \centering
    \includegraphics[width=\linewidth]{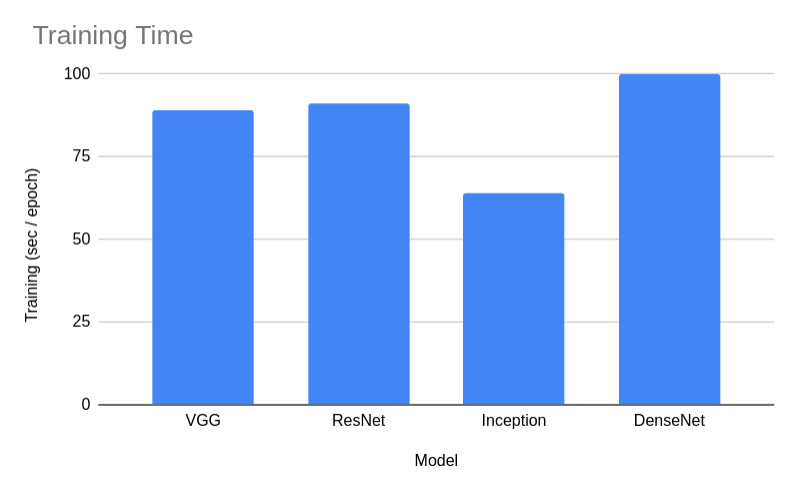}
    \caption{\label{figure:train_time}Training time per epoch for each model.}
    \vspace{-6mm}
\end{figure}

To avoid averaging the scanpaths using this function, a single scanpath was randomly chosen per image for training. %This was possible because of the size of available data for training.     
The Adam optimizer was used with a learning rate of 0.0003. The training was done for 25 epochs on each model. The training was conducted using a single Nvidia Quadro P5000 GPU. All of our models took less than 100 seconds per epoch during training as shown in Fig.\ref{figure:train_time}, while PathGAN used 6 Nvidia K80 for 72 hours for their training. This shows the benefit of using a fully convolutional architectures on the acceleration of training time.

\section{3. Experimental results}
\label{expiriments}
\begin{figure*}[ht!]
    \vskip 0.5cm
    \includegraphics[width=\textwidth]{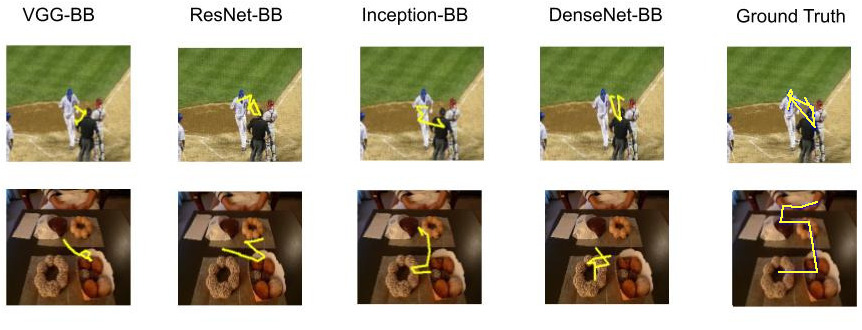}
    \caption{\label{scanpath_viz}Visualization of the predicted scanpaths of the considered models for some test images of Salicon dataset.}
    \vspace{-6mm}
\end{figure*}

In this section, we present first qualitative results, following by some quantitative results using some well-known metrics.

\subsection{3.1. Datasets}

In order to evaluate the efficiency of the proposed approach, two well-known datasets have been used for this study.

\textbf{Salicon \cite{Salicon}:} This dataset is proposed as part of the Salicon competition challenge where the goal was to predict the saliency maps of 2D images. We used 10000 images for our training and validation with a $90\%-10\%$  train-validation split, and $5000$ images for testing. Each image has corresponding saliency map and scanpaths. %Fig. \ref{figure_Salicon} shows a sample of images of this dataset as well as a corresponding scanpath.
%It is composed of 10000 images for the training with a 90\%-10\% train-validation split and 5000 images for the test. For each image, the scanpath and the corresponding saliency map is given. Fig. \ref{figure_Salicon} shows a sample of images of this dataset as well as a corresponding scanpath.

%\begin{figure}
%    \vskip 0.5cm
%    \includegraphics[width=\linewidth]{figure/salicon_viz.jpg}
%    \caption{\label{figure_Salicon}Sample of images and the corresponding scanpaths from Salicon dataset}
%\end{figure}

\textbf{MIT1003 \cite{mit1003}:} This dataset is usually used in conjunction with the MIT300 dataset for their competition. It consists of $1003$ natural images with their saliency maps and scanpaths. The dataset was used as a whole for a cross-dataset neutral comparison between our proposed methods and the other models. %Fig. \ref{figure_MIT} shows some images of the MIT1003 dataset as well as the corresponding scanpaths.

%\begin{figure}
%    \vskip 0.5cm
%    \includegraphics[width=\linewidth]{figure/mit_viz.jpg}
%    \caption{\label{figure_MIT}Sample of images and the corresponding scanpaths from MIT 1003 dataset.}
%\end{figure}

\subsection{3.2. Qualitative results}

\begin{figure}
    \centering
    \vskip 0.5cm
    \includegraphics[width=\linewidth]{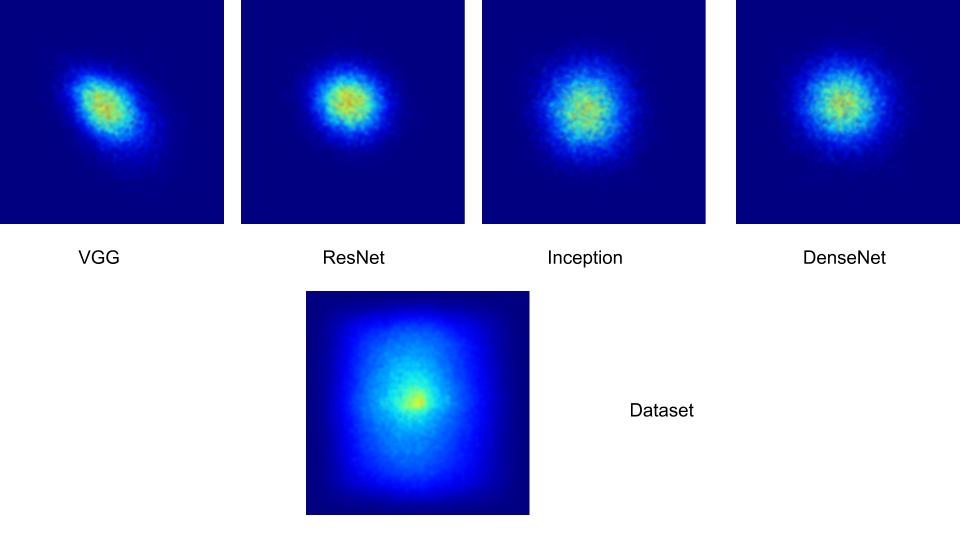}
    
    \caption{\label{distribution}Comparison of the distribution of the predicted and the ground truth scanpaths for each model on Salicon dataset.}
    \vspace{-6mm}
\end{figure}

Fig.\ref{scanpath_viz} shows predicted scanpaths for some test images of Salicon dataset. These results demonstrate the concordance of the scanpaths with the salient regions and the semantics content of the image. We observed that VGG-BB scanpaths have more simplistic and centered shapes, while Inception-BB scanpaths look more complex in shape and stretch on a bigger area of interest the image. Whereas, ResNet-BB and DenseNet-BB scanpaths have the most complex shapes and it is noted that a lot of scanpaths predicted by these models has crossing points (i.e sometimes the scanpath returns a previous region or crosses an area that was already explored). 
We also notice that the saccades lengths are a little shorter compared to  the ground truth, this is probably due to  the limitations of the MSE loss function used during training which amplify the central bias of the data Fig.\ref{distribution}.

Fig.\ref{distribution} shows the density distribution of the predicted fixation points for each model. For comparison, we also show the density distribution of the ground truth scanpath fixations of the validation set of Salicon dataset. We noticed that all the considered models could successfully find the central bias of the dataset. However, the models still have some limitations as the distributions were highly concentrated and did not completely fit the dataset distribution on the edges of the images. It is worth noting that the Inception-BB covers a bigger area than other models, meaning it has higher deviation compared to other models, which is further emphasised by the results in Fig. \ref{scanpath_viz}.    

\subsection{3.3. Quantitative analysis}

\subsubsection{3.3.1. Evaluation metrics}

In this work, 3 different metrics were used to evaluate and compare the results of each considered model: 2 hybrid metrics (i.e. NSS \cite{NSS} and Congruency \cite{congruency}) and 1 metric dedicated to scanpath comparison (i.e. multimatch \cite{multimatch}). The former use saliency maps and scanpaths, while the latter is a vector-based method that compares 2 scanpaths. Each on these metrics is briefly described in this section.
%\vspace{-3mm}
\begin{enumerate}
    \item 
\textbf{NSS} (Normalized Scanpath Saliency) \cite{NSS} is a hybrid metric used to compare saliency maps and predicted scanpaths. %A binary fixation map is first constructed using the predicted scanpath. The map is then multiplied pixel-wise with a normalized saliency map (i.e. zero mean and one unit standard deviation). 
The arithmetic mean of the scanpath fixations saliency values is calculated for the NSS score as follows: %of the resulting multiplication is calculated for the NSS score, which provides the mean saliency value of the fixations points of the scanpath. NSS score is given as follows:
\begin{equation}
    NSS=\frac{1}{N} \sum (P*Q_b) \quad  
    \label{eq:NSS-pre}
\end{equation}
with 
\begin{equation}
    \quad P=\frac{S - \mu(S)}{\sigma(S)} 
    \label{eq:NSS}
\end{equation}
where \textbf{Q\textsubscript{b}} is the binary fixation map derived from the scanpath. 
\textbf{S} is the saliency map and \textbf{P} is the normalized saliency map. $\bm{\mu}$ and $\bm{\sigma}$ are the mean and the standard deviation, respectively.
\newline

 \item 
\textbf{Congruency} \cite{congruency} is a hybrid metric that measures the dispersion of the scanpaths, and the variability of a scanpath in accordance to the others. %Fig. \ref{figure:congruency} summarizes the principal steps of this metric.
The ground truth saliency map is dynamically thresholded to binary map using the Otsu method, which allows to separate the salient and no-salient regions. The congruency is the ratio of fixation points from the scanpath that belong to the salient regions of the binary map. It is computed as follows: 
\begin{equation}
    C=\frac{1}{N} \sum_{i=1}^{N} (BinMap_i * FixMap_i) 
    \label{eq:congruency}
\end{equation}
where $BinMap_i$ is the thresholded binary map, and $FixMap_i$ is the fixation map for the predicted scanpath.   \\

Unlike NSS, this metric does not take into account the salient value of a fixation point, and thus all salient points are treated equally.\\

%A fixation map is first constructed from the ground truth scanpaths and passed through a convolution with a Gaussian kernel in order to generate a saliency map for the image. The latter map is then dynamically thresholded using the Otsu's method \cite{otsu}, which allows to separate the salient and no-salient regions. The congruency is finally calculated by the ratio of fixation points that are in the salient region of the binary map to the number of fixation points in the scanpath as follows:  

%\[C=\frac{1}{N} \sum_{i=1}^{N} (Map_i-Map_i) \]
%where ...

%Unlike NSS, this metric does not take into account the salient value of a fixation point, and thus all salient points are treated equally.
%\newline

\item
\textbf{MultiMatch} (Jarodzka) was introduced by Jarodzka et al. \cite{multimatch}. This metric takes the scanpaths as multi-dimensional vectors and compares their similarity based on 5 different characteristics:

\begin{enumerate}
  \item \textbf{Shape:} difference in shape between scanpaths $(\overline{u}  - \overline{v})$.
  \item \textbf{Direction:} difference in angle between saccades.
  \item \textbf{Length:} difference in length between saccades.
  \item \textbf{Position:} difference in position between fixations positions.
  \item \textbf{Duration:} difference in saccades duration.
\end{enumerate}

%Shape, Length and Position are normalized to the screen diagonal, while Direction is normalized to $\Pi$ and the Duration is normalized to the longest duration. \textit{Score} representing complete 
The overall score was here calculated using the arithmetic mean of the first 4 characteristics following equation \ref{eq:MM}, excluding the Duration since our models don’t predict the timestamps of the fixations. \\
\begin{equation}
    Score=\frac{Shape+Direction+Length+Position}{4}
    \label{eq:MM}
\end{equation}
%\[ Score=\frac{Shape+Direction+Length+Position}{4}\]

\end{enumerate}

\subsubsection{3.3.2. Evaluation}

Our models were compared with several other methods of the state-of-the-art. They were tested on 5000 images from the Salicon dataset which were not used during training. In order to validate the generalization ability of our method, our models were also evaluated on MIT1003 dataset without any further fine-tuning for cross-dataset testing.

%Table \ref{tab:NSS_test} shows the results obtained in terms of NSS metric. ResNet-BB scored the highest on both Salicon and MIT1003. The second highest score was achieved by Inception-BB on both datasets. On Salicon, Le Meur scored the third highest, significantly  higher than the rest of the models. On MIT1003, both VGG-BB and DenseNet-BB achieved a higher score than Le Meur. PathGAN achieved the lowest score On both datasets. The latter result is in accordance with the sparser distribution of spatial locations shown in \cite{pathgan}.

Results for the Salicon \cite{Salicon} dataset were presented in Table 2. While the results for MIT1003\cite{mit1003} dataset are  shown in Table 3. 

ResNet-BB and Inception-BB achieved the highest results on the NSS metric for both datasets. While VGG-BB and DenseNet-BB achieved competitive results with rest of the models. We also notice that Le Meur\cite{lemeur} and G-Eymol achieved high results compared to the other models. For Congruency, Le Meur \cite{lemeur} realized the best score on both datasets. SalyPath achieved high comparable results on the Salicon dataset. G-Eymol also achieved high results on both datasets. These models did not generalize well on the MIT1003 dataset for this metric, as we notice a steep decline in performance. Our models displayed more consistent results across datasets because the decline in results was small. 

Our models along with the SalyPath network demonstrate the highest results on the shape component of the MultiMatch metric, followed by the rest of the models. ResNet-BB and Inception-BB tied for the best performance, while Le Meur and G-Eymol achieved the lowest performance. DenseNet-BB achieved the best results for the direction and length components on  Salicon, closely followed by ResNet-BB and Inception-BB. Inception-BB had the best performance on the MIT1003 dataset for the Direction metric followed by SalyPath and DenseNet-BB. VGG-BB accomplished the best performance for the position component on Salicon, followed by DenseNet-BB, but their results dropped on the MIT1003 dataset, in which DSCM (ResNet) headed the scores. The best mean score was achieved by DenseNet-BB on the Salicon dataset, followed by SalyPath. For the MIT1003 dataset, we also reported the results found by \cite{DCSM} on the MultiMatch metric, we couldn't do the same for Salicon model as their model was not provided for testing.

%Over all Inception-BB and ResNet-BB  achieved the highest score for the hybrid metrics, closely followed by DenseNet-BB and ResNet-BB respectively, Le Meur\cite{lemeur} scored better then VGG-BB , Both the DCSM models and PathGAN\cite{pathgan}. 

%Table \ref{tab:congruency_results} shows the results obtained for congruency on both datasets. Le Meur\cite{lemeur} achieved the highest score on both datasets, surpassing the deep-based models by a wide margin on the Salicon dataset. ResNet-BB achieved the second highest score on both datasets, while the third highest score was reached by Inception-BB on Salicon, followed by VGG-BB, DenseNet-BB and PathGAN. On the MIT1003 dataset, the third  highest score was obtained by DenseNet-BB, followed respectively by Inception-BB, VGG-BB and PathGAN.

\begin{table*}[ht]
\begin{center}
\begin{tabular}{ c c c c c c c c }
\hline
\textbf{Model} & \textbf{Shape } & \textbf{Direction} & \textbf{Length} & \textbf{Position} & \textbf{MM Score} & \textbf{NSS} & \textbf{Congruency}\\ 
\hline
 PathGAN & 0.9608 & 0.5698 & 0.9530 & 0.8172 & 0.8252  & -0.2904  &  0.0825 \\ 
 \hline
 Le Meur & 0.9505 & 0.6231 & 0.9488 & 0.8605 & 0.8457 &  0.8780 &   \textbf{0.4784}   \\ 
 \hline
 G-Eymol & 0.9338 & 0.6271 & 0.9521 & 0.8967 &  0,8524 &  0.8727 & 0.3449 \\
 \hline
 SALYPATH  & 0.9659  & 0.6275 & 0.9521 & 0.8965  &   0,8605  & 0.3472 &  0.4572     \\ 
 \hline
 VGG-BB & 0.9701 & 0.6001 & 0.9471 & \textbf{0.9098} &  0.8568  & 0.6848 & 0.1499  \\
 \hline
 ResNet-BB  & \textbf{0.9705} & 0.6395 & 0.9532 & 0.8230 & 0.8465 & \textbf{1.1921} & 0.1532\\
 \hline
 Inception-BB  & \textbf{0.9705} & 0.6384 & 0.9534 & 0.8233 & 0.8464 & 1.0606 &  0.1517 \\
 \hline
 DenseNet-BB  & 0.9673 & \textbf{0.6584} & \textbf{0.9608} & 0.9081 & \textbf{0.8737} & 0.5939 &  0.1449 \\
 \hline
 
\end{tabular}
\label{tab:MM-salicon}
\caption{ Table 2. Results of models for MultiMatch $\uparrow$ on the Salicon dataset}
\end{center}
\end{table*}

%Table \ref{tab:MM-salicon} shows the results obtained on the Salicon dataset using MultiMatch. DenseNet-BB achieved the highest mean score, followed by VGG-BB. The third place was reached by ResNet-BB, closely followed by Inception-BB, then Le Meur and lastly PathGAN. All our models obtained a higher mean scores than both Le Meur and PathGAN. For the shape character, the best score was shared by ResNet-BB and Inception-BB, closely followed by VGG-BB.
%For direction, DenseNet-BB achieved the best score, followed by ResNet-B and Inception-BB. Le Meur ranked higher than VGG-BB and PathGAN. 

%For Leangth, DenseNet-BB ranked the highest followed by Inception-BB and ResNet-BB. Le Meur achieved a higher score than VGG-BB and PathGAN.

%VGG-BB had the best results for the position characteristic followed by DenseNet-BB then Le Meur achieved the third highest score, then came Inception-BB and ResNet-BB respectively, finally PathGAN ranked last.      

\begin{table*}[ht]
\begin{center}
\begin{tabular}{ c c c c c c c c}
\hline
\textbf{Model} & \textbf{Shape } & \textbf{Direction} & \textbf{Length} & \textbf{Position} & \textbf{MM score} & \textbf{NSS} & \textbf{Congruency}\\ 
\hline
 PathGAN & 0.9237 & 0.5630 & 0.8929 & 0.8124 & 0.7561   & -0.2750  &  0.0209\\ 
 \hline
 Le Meur & 0.9241  &  0.6378  & \textbf{0.9171}  & 0.7749 & 0,8135 & 0.8508   &  \textbf{0.1974} \\ 
 \hline
 G-Eymol & 0.8885 & 0.5954 & 0.8580 & 0.7800 &  0,7805 & 0.8700 & 0.1105 \\
 \hline
 DCSM (VGG)  & 0.8720 & 0.6420 & 0.8730 & 0.8160 & 0,8007 & - & - \\
 \hline
 DCSM (ResNet) & 0.8780 & 0.5890 & 0.8580 & \textbf{0.8220} &  0,7868 & - & -\\
 \hline
 SALYPATH  & 0.9363  & 0.6507 & 0.9046 & 0.7983  &   \textbf{0,8225} &  0.1595  &   0.0916    \\ 
 \hline
 VGG-BB & 0.9350 & 0.6004 & 0.8887 & 0.7839 &  0.8020 & 0.8979 &   0.1276\\
 \hline
 ResNet-BB  & \textbf{0.9373} & 0.6450 & 0.9034 & 0.7816 & 0.8168  & \textbf{1.0727}  & 0.1626 \\
 \hline
 Inception-BB  & 0.9372 & \textbf{0.6537} & 0.9041 & 0.7793 & 0.8186  &  0.9155 &   0.1464 \\
 \hline
 DenseNet-BB  & 0.9371  & 0.6489 & 0.9056  & 0.7812 & 0.8182 &  0.8871 & 0.1581  \\
 
 \hline

\end{tabular}

\caption{\label{tab:MM-MIT} Table 3. Results of models for the MultiMatch $\uparrow$ metric on the MIT1003 dataset}
\end{center}
\end{table*}

%Table\ref{tab:MM-MIT} shows the results obtained on the MIT1003\cite{Judd} dataset using MultiMatch\cite{multimatch}. 

%For the MIT1003 dataset we also reported the results found by \cite{DCSM} on the MultiMatch metric, we couldn't do the same for Salicon model as their model was not provided for testing.   

%ResNet-BB achieved the best result on the shape characteristic followed closely by Inception-BB and DenseNet-BB then VGG-BB, Le Meur\cite{lemeur} and PathGAN\cite{pathgan}, finally the worst results were achieved by the 2 DCSM\cite{DCSM} models.      

%On the Direction component Inception-BB scored the highest followed by DenseNet-BB,ResNet-BB and DCSM(VGG)\cite{DCSM} then Le Meur\cite{lemeur}, VGG-BB, DCSM(ResNet)\cite{DCSM} and finally PathGAN\cite{pathgan}.      

%For the characteristic of Length, Le Meur scored the highest then DenseNet-BB, Inception-BB respectively and ResNet-BB, followed by PathGAN\cite{pathgan} , Le Meur\cite{lemeur} and thee 2 DCSM \cite{DCSM} models. 

%Lastly for the Position component, DCSM (ResNet) and DCSM (VGG) \cite{DCSM} scored the the highest respectively followed by PathGAN\cite{pathgan} ,  VGG-BB,ResNet-BB,DenseNet-BB then Inception-BB and Le Meur\cite{lemeur}.     

\section{4. Discussion}
\label{discussion}

We noticed that Le Meur and G-Eymol were able to achieve higher scores than some deep learning models on hybrid metrics (NSS and Congruency), surpassing VGG-BB on Salicon\cite{Salicon}, or achieving the very high scores for Congruency on both datasets. This is due to their procedure of generating scanpaths, as they select the fixation points from a predicted saliency map. In the case of Le meur \cite{lemeur}, they select the points based the highest saliency values after among other criteria, while G-Eymol is more likely to be attracted to salient region due to the inherit characteristic of the gravity model.

Out of our models, ResNet-BB achieved the highest score for the hayride metrics. This suggests the beneficial effect of propagating the image low level features deeper in the network for the task of scanpath prediction. DenseNet-BB is another model  which uses skip connections and has achieved the best results for the MultiMatch metric. This also reinforces the importance of using skip connections to  propagate lower level features. Both the qualitative an quantitative results for Inception-BB network results, suggest that it is the most useful for visual attention tasks. This might be due to the effects of using multiple scales for convolution kernels in a parallel manner, similar to the use of multi-scale data in many visual attention models in the past like \cite{Itti} and \cite{Salicon}. These previous results highly accentuates the usefulness of multi-scale and multi-level dense data representations using local skip connections in improving the task of scanpath prediction. Our results also suggest that fine-tuning simple pre-trained  models using uncomplicated and straight forward training procedures can yield competitive results that can surpass state-of-the-art models for certain criteria.

\section{5. Conclusion}
\label{conclusion}

In this work we proposed a simple fully convolutional architecture to predict visual scanpaths on natural images. By leveraging the capabilities of commun computer vision models like (VGG, ResNet, etc.), and transforming their outputs using a convolutional layer that outputs a feature map representing the predicted scanpath. Through  experiments on 2 different datasets (i.e. Salicon and MIT1003), we compared the performance of our proposed architectures with other benchmark models, and achieved very competitive results on different quantitative metrics. we also displayed their ability to successfully generalize to other data distributions. Through qualitative visualizations, we demonstrated their ability to generate plausible and relevant scanpaths. And by comparing the visual distributions of the scanpaths, we demonstrated the ability of detecting the central bias by the our models. 

%We used 3 metrics to compare the models, with 2 hybrid  metrics (NSS, Congruency) and a vector based metric (MultiMatch ).
%Our Models achieved state of the art on some metrics and very  competitive results on the others.   

This study  allowed us to compare the different backbone architectures, and the usefulness of their characteristics for this task. It also demonstrates that using simple architectures and simple training procedures can be as effective as using complex frameworks, and training protocols.


\small

\begin{thebibliography}{00}

\bibitem{feature_integration}Treisman, Anne M., and Garry Gelade., \emph{A feature-integration theory of attention.} Cognitive psychology 12.1 (1980): 97-136.

\bibitem{QAChetouaniICIP2018}A. Chetouani, "Convolutional Neural Network and Saliency Selection for Blind Image Quality Assessment," 2018 25th IEEE International Conference on Image Processing (ICIP), 2018, pp. 2835-2839, doi: 10.1109/ICIP.2018.8451654.

\bibitem{saliencyComp1}Patel, Yash, Srikar Appalaraju, and R. Manmatha. "Saliency Driven Perceptual Image Compression." Proceedings of the IEEE/CVF Winter Conference on Applications of Computer Vision. 2021.

%\bibitem{saliencyComp2}Stella, X. Yu, and Dimitri A. Lisin. "Image compression based on visual saliency at individual scales." International Symposium on Visual Computing. Springer, Berlin, Heidelberg, 2009.

\bibitem{saliencyCapt1}M. Cornia, L. Baraldi, G. Serra and R. Cucchiara, "Visual saliency for image captioning in new multimedia services," 2017 IEEE International Conference on Multimedia \& Expo Workshops (ICMEW), 2017, pp. 309-314, doi: 10.1109/ICMEW.2017.8026277.

\bibitem{retrivalSal}Wang, Haoxiang, et al. "Visual saliency guided complex image retrieval." Pattern Recognition Letters 130 (2020): 64-72.

\bibitem{KochUllman}Koch, C. and S. Ullman. “Shifts in selective visual attention: towards the underlying neural circuitry.” Human neurobiology 4 4 (1985): 219-27 .
%\bibitem{saliencyCapt2}L. Zhou, Y. Zhang, Y. -G. Jiang, T. Zhang and W. Fan, "Re-Caption: Saliency-Enhanced Image Captioning Through Two-Phase Learning," in IEEE Transactions on Image Processing, vol. 29, pp. 694-709, 2020, doi: 10.1109/TIP.2019.2928144.

\bibitem{Itti} Itti, Laurent, and Christof Koch. "Computational modelling of visual attention." Nature reviews neuroscience 2.3 (2001): 194-203.

\bibitem{GBVS} Harel, Jonathan, Christof Koch, and Pietro Perona. "Graph-based visual saliency." Advances in neural information processing systems. 2007.

\bibitem{infosal} Bruce, N. and Tsotsos, J.K. (2005) Saliency based on information maximization. Advances in Neural Information Processing Systems, 18, 155–162.

\bibitem{spectral1} Peters, R.J. and Itti, L. (2008) The role of Fourier phase information in predicting saliency. Proceedings of Vision Science Society Annual Meeting (VSS08).

\bibitem{spectral2} Guo, C.L., Ma, Q. and Zhang, L.M. (2008) Spatio-temporal saliency detection using phase spectrum of
quaternion Fourier transform. Proceedings of IEEE Conference on Computer Vision and Pattern Recognition (CVPR2008).

%\bibitem{Sternberg} Sternberg, R.J., Sternberg, K., Mio, J.: Cognitive Psychology, 6th edn. Cengage Learning (2008)

%\bibitem{Raichle} Raichle, M.E.: The brain’s dark energy. Scientific American Magazine, 44–49 (2010)

\bibitem{Judd}Judd, Tilke, Frédo Durand, and Antonio Torralba. "A benchmark of computational models of saliency to predict human fixations." (2012).

%\bibitem{turckergaze} Xu, Pingmei, et al. "Turkergaze: Crowdsourcing saliency with webcam based eye tracking." arXiv preprint arXiv:1504.06755 (2015).

\bibitem{Salicon} Jiang, Ming, et al. "Salicon: Saliency in context." Proceedings of the IEEE conference on computer vision and pattern recognition. 2015.

\bibitem{PanDSN} Pan, Junting, et al. "Shallow and deep convolutional networks for saliency prediction." Proceedings of the IEEE conference on computer vision and pattern recognition. 2016.

\bibitem{mlnet} Cornia, Marcella, et al. "A deep multi-level network for saliency prediction." 2016 23rd International Conference on Pattern Recognition (ICPR). IEEE, 2016.

\bibitem{salgan} Pan, Junting, et al. "Salgan: Visual saliency prediction with generative adversarial networks." arXiv preprint arXiv:1701.01081 (2017).

\bibitem{sam} Cornia, Marcella, et al. "Predicting human eye fixations via an lstm-based saliency attentive model." IEEE Transactions on Image Processing 27.10 (2018): 5142-5154.

\bibitem{unisal} Droste, Richard, Jianbo Jiao, and J. Alison Noble. "Unified Image and Video Saliency Modeling." arXiv preprint arXiv:2003.05477 (2020).

%\bibitem{emlnet} Jia, Sen. “EML-NET: An Expandable Multi-Layer NETwork for Saliency Prediction.” Image Vis. Comput. 95 (2020): 103887.


%\bibitem{gazegan} Z. Che, A. Borji, G. Zhai, X. Min, G. Guo and P. Le Callet, "How is Gaze Influenced by Image Transformations? Dataset and Model," in IEEE Transactions on Image Processing, vol. 29, pp. 2287-2300, 2020, doi: 10.1109/TIP.2019.2945857.

\bibitem{boccignone} Giuseppe Boccignone, Mario Ferraro,Modelling gaze shift as a constrained random walk,Physica A: Statistical Mechanics and its Applications,Volume 331, Issues 1–2,2004,Pages 207-218

\bibitem{cerf} Cerf, Moran \& Harel, Jonathan \& Einhäuser, Wolfgang \& Koch, Christof. (2007). Predicting human gaze using low-level saliency combined with face detection. Adv Neural Inf Process Syst. 20. 

\bibitem{lemeur}Le Meur, Olivier \& Liu, Zhi. (2015). Saccadic model of eye movements for free-viewing condition. Vision research. 116. 10.1016/j.visres.2014.12.026. 

%\bibitem{LSPI} Jiang, Ming, et al. "Learning to predict sequences of human visual fixations." IEEE transactions on neural networks and learning systems 27.6 (2016): 1241-1252.

\bibitem{shao} Shao, Xuan, et al. "Scanpath prediction based on high-level features and memory bias." International Conference on Neural Information Processing. Springer, Cham, 2017.

\bibitem{saltinet} Assens Reina, Marc, et al. "Saltinet: Scan-path prediction on 360 degree images using saliency volumes." Proceedings of the IEEE International Conference on Computer Vision Workshops. 2017.

\bibitem{pathgan} Assens, Marc, et al. "PathGAN: visual scanpath prediction with generative adversarial networks." Proceedings of the European Conference on Computer Vision (ECCV). 2018.

%\bibitem{extdeepgaze} Kümmerer, Matthias & Wallis, Thomas & Bethge, Matthias. (2018). Extending DeepGaze II: Scanpath prediction from deep features. Journal of Vision. 18. 371. 10.1167/18.10.371. 

\bibitem{deepgaze1ref} Krizhevsky, Alex, Sutskever, Ilya, and Hinton, Geoffrey E. Imagenet classification with deep convolutional neural networks. In Advances in neural information processing systems, pp. 1097–1105, 2012.

\bibitem{deepgaze1} Kümmerer, Matthias, Lucas Theis, and Matthias Bethge. "Deep gaze i: Boosting saliency prediction with feature maps trained on imagenet." arXiv preprint arXiv:1411.1045 (2014).

\bibitem{deepgaze2} Kümmerer, Matthias, Thomas SA Wallis, and Matthias Bethge. "DeepGaze II: Reading fixations from deep features trained on object recognition." arXiv preprint arXiv:1610.01563 (2016).

\bibitem{HMMLSTMConv}Verma, Ashish, and Debashis Sen. "HMM-based Convolutional LSTM for Visual Scanpath Prediction." 2019 27th European Signal Processing Conference (EUSIPCO). IEEE, 2019.

\bibitem{G-Eymol}Zanca, Dario, Stefano Melacci, and Marco Gori. "Gravitational laws of focus of attention." IEEE transactions on pattern analysis and machine intelligence 42.12 (2019): 2983-2995.

\bibitem{DCSM}Bao, Wentao, and Zhenzhong Chen. "Human Scanpath Prediction based on Deep Convolutional Saccadic Model." Neurocomputing (2020).

\bibitem{salypath}M. A. Kerkouri, M. Tliba, A. Chetouani and R. Harba, "Salypath: A Deep-Based Architecture For Visual Attention Prediction," 2021 IEEE International Conference on Image Processing (ICIP), 2021, pp. 1464-1468, doi: 10.1109/ICIP42928.2021.9506295.

\bibitem{mit1003} Judd, Tilke, et al. "Learning to predict where humans look." 2009 IEEE 12th international conference on computer vision. IEEE, 2009.

\bibitem{vgg} Simonyan, Karen, and Andrew Zisserman. "Very deep convolutional networks for large-scale image recognition." arXiv preprint arXiv:1409.1556 (2014).

\bibitem{resnet} He, Kaiming, et al. "Deep residual learning for image recognition." Proceedings of the IEEE conference on computer vision and pattern recognition. 2016.

\bibitem{inception} Szegedy, Christian, et al. "Rethinking the inception architecture for computer vision." Proceedings of the IEEE conference on computer vision and pattern recognition. 2016.

\bibitem{densenet} Iandola, Forrest, et al. "Densenet: Implementing efficient convnet descriptor pyramids." arXiv preprint arXiv:1404.1869 (2014).

%\bibitem{adam} Kingma, Diederik P., and Jimmy Ba. "Adam: A method for stochastic optimization." arXiv preprint arXiv:1412.6980 (2014).

%\bibitem{imagenet_org} Deng, Jia, et al. "Imagenet: A large-scale hierarchical image database." 2009 IEEE conference on computer vision and pattern recognition. Ieee, 2009.

%\bibitem{imagenet2015} Russakovsky, Olga, et al. "Imagenet large scale visual recognition challenge." International journal of computer vision 115.3 (2015): 211-252.

\bibitem{multimatch} Halszka Jarodzka, Kenneth Holmqvist, and Marcus Nyström. 2010. A vector-based, multidimensional scanpath similarity measure. In Proceedings of the 2010 Symposium on Eye-Tracking Research \& Applications (ETRA '10). Association for Computing Machinery, New York, NY, USA, 211–218.

\bibitem{congruency} Le Meur, Olivier \& Baccino, Thierry \& Roumy, A.. (2011). Prediction of the Inter-Observer Visual Congruency (IOVC) and Application to Image Ranking. Proceedings of ACM Multimedia. 373-382. 10.1145/2072298.2072347. 

\bibitem{NSS} Peters, Robert J., et al. "Components of bottom-up gaze allocation in natural images." Vision research 45.18 (2005): 2397-2416.

%\bibitem{otsu} N. Otsu, "A Threshold Selection Method from Gray-Level Histograms," in IEEE Transactions on Systems, Man, and Cybernetics, vol. 9, no. 1, pp. 62-66, Jan. 1979, doi: 10.1109/TSMC.1979.4310076.
 
\end{thebibliography}
\end{document}